\documentclass[journal]{IEEEtran}

\ifCLASSINFOpdf
\usepackage[pdftex]{graphicx}
\else
\fi

\usepackage{amsmath}
\usepackage{amsfonts}
\usepackage{graphicx,verbatim}
\usepackage{tabularx}
\usepackage{multirow}
\usepackage{booktabs}
\usepackage{amssymb}

\usepackage{hyperref} 
\usepackage{xcolor}

\newcommand{\code}[1]{\texttt{#1}}
\DeclareSymbolFont{bbold}{U}{bbold}{m}{n}
\DeclareSymbolFontAlphabet{\mathbbold}{bbold}

\begin{document}
%
\title{Point Tracking in Surgery--The 2024 Surgical Tattoos in Infrared (STIR) Challenge}
%
%
%

\author{Adam Schmidt$^{1,*}$, 
    Mert Asim Karaoglu$^{2, 3}$, 
    Soham Sinha$^{4}$,
    Mingang Jang$^{5}$,
    Ho-Gun Ha$^{5}$,
    Kyungmin Jung$^{5}$,
    Kyeongmo Gu$^{5}$,
    Ihsan Ullah$^{5}$, 
    Hyunki Lee$^{5}$,
    Jonáš Šerých$^{6}$,
    Michal Neoral$^{6}$,
    Jiří Matas$^{6}$,
    Rulin Zhou$^{7,9}$, 
    Wenlong He$^9$, 
    An Wang$^8$, 
    Hongliang Ren$^{7,8}$, 
    Bruno Silva$^{10, 11, 12}$, 
    Sandro Queirós$^{10, 11}$, 
    Estêvão Lima$^{10, 11}$, 
    João L. Vilaça$^{12, 13}$, 
    Shunsuke Kikuchi$^{14,15}$,
    Atsushi Kouno$^{14}$,
    Hiroki Matsuzaki$^{14}$,
    Tongtong Li$^{16}$, 
    Yulu Chen$^{16}$,
    Ling Li$^{16,17}$,
    Xiang Ma$^{16}$,
    Xiaojian Li$^{16,17}$,
    Mona Sheikh Zeinoddin$^{18}$, 
    Xu Wang$^{18}$, 
    Zafer Tandogdu$^{19}$, 
    Greg Shaw$^{19}$, 
    Evangelos Mazomenos $^{18}$, 
    Danail Stoyanov $^{18}$, 
    Yuxin Chen$^{20}$,
    Zijian Wu$^{20}$,
    Alexander Ladikos$^2$, 
    Simon DiMaio$^1$, 
    Septimiu E. Salcudean$^{20}$, %
    Omid Mohareri$^1$ 

\thanks{$^1$ Intuitive Surgical Inc., Sunnyvale, USA,
$^2$ ImFusion GmbH, Munich, Germany,
$^3$ Technical University of Munich, Munich, Germany,
$^4$ NVIDIA, Santa Clara, USA}
\thanks{$^{5}$ Division of Intelligent Robotics, Daegu Gyeongbuk Institute of Science and Technology (DGIST), South Korea}
\thanks{$^{6}$ CMP Visual Recognition Group, Faculty of Electrical Engineering, Czech Technical University in Prague}
\thanks{$^7$ Shenzhen Research Institute, China,
$^8$ The Chinese University of Hong Kong, China,
$^9$ Shenzhen University, China}
\thanks{$^{10}$ Life and Health Sciences Research Institute (ICVS), School of Medicine, University of Minho, Braga, Portugal,
$^{11}$ ICVS/3B’s - PT Government Associate Laboratory, Braga/Guimarães, Portugal,
$^{12}$ 2Ai –School of Technology, IPCA, Barcelos, Portugal
$^{13}$ LASI – Associate Laboratory of Intelligent Systems, Guimarães, Portugal}
\thanks{$^{14}$ Jmees Inc, Japan,
$^{15}$ University of California - Los Angeles (UCLA), USA}
\thanks{$^{16}$ School of Management, Hefei University of Technology, Hefei 230009, China,
$^{17}$ Key Laboratory of Process Optimization and Intelligent Decision-making, Ministry of Education, Hefei, China.} 
\thanks{$^{18}$ Hawkes Institute, University College London, London, UK,
$^{19}$ Dept of Urology, University College London Hospitals, UK}
\thanks{$^{20}$ University of British Columbia}
\thanks{$^{*}$ Corresponding author: {\tt\small adam.schmidt at intusurg.com}}

}

%
%

\markboth{Point Tracking in Surgery--The 2024 Surgical Tattoos in Infrared (STIR) Challenge}%
{Point Tracking in Surgery--The 2024 Surgical Tattoos in Infrared (STIR) Challenge}
%


\maketitle

\begin{abstract}
Understanding tissue motion in surgery is crucial to enable applications in downstream tasks such as segmentation, 3D reconstruction, virtual tissue landmarking, autonomous probe-based scanning, and subtask autonomy.
Labeled data are essential to enabling algorithms in these downstream tasks since they allow us to quantify and train algorithms.
This paper introduces a point tracking challenge to address this, wherein participants can submit their algorithms for quantification.
The submitted algorithms are evaluated using a dataset named surgical tattoos in infrared (STIR), with the challenge aptly named the STIR Challenge 2024.
The STIR Challenge 2024 comprises two quantitative components: accuracy and efficiency.
The accuracy component tests the accuracy of algorithms on \em{in vivo} and \em{ex vivo} sequences.
The efficiency component tests the latency of algorithm inference.
The challenge was conducted as a part of MICCAI EndoVis 2024.
In this challenge, we had 8 total teams, with 4 teams submitting before and 4 submitting after challenge day.
This paper details the STIR Challenge 2024, which serves to move the field towards more accurate and efficient algorithms for spatial understanding in surgery.
In this paper we summarize the design, submissions, and results from the challenge.
The challenge dataset is available here: \url{https://zenodo.org/records/14803158}, and the code for baseline models and metric calculation is available here: \url{https://github.com/athaddius/STIRMetrics}
\end{abstract}


\IEEEpeerreviewmaketitle

\section{Introduction}

\IEEEPARstart{T}{he} 2024 STIR challenge is designed to help improve tracking and reconstruction methods in surgery.
Knowledge of tissue motion and location is critical to enable many tasks in medical computer vision~\cite{schmidtTrackingMappingMedical2024a}.
Improved accuracy of motion estimation is essential to enable automated dexterity~\cite{kamAutonomousSystemVaginal2023}, autonomous scanning~\cite{zhan2020autonomous}, and virtual landmarking.
Improved performance here will likely also benefit foundation models, where physical priors can be incorporated into pretraining.
This challenge marks the first in kind for point tracking, wherein we use infrared labels to quantify the performance of submitted methods.
The data used in the challenge comprises 60 sequences with each sequence including an average of 8 points.

\begin{figure*}[t]
	\centering
	\includegraphics[width=\textwidth]{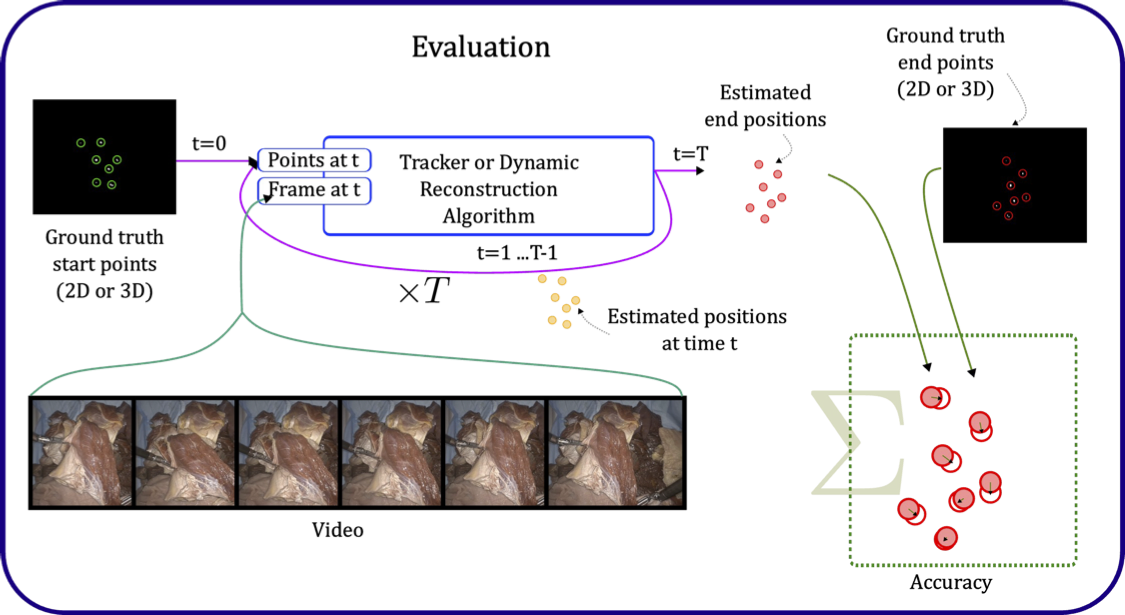}
	\caption{This figure describes the STIR Challenge 2024. Participants submit their algorithms in a docker container. The algorithm receives a video and a list of start points from each sequence in the dataset. Participants use their tracker to estimate the motion of a set of points for every frame in a video. Videos are provided in a streaming manner. The final estimates are then compared to the ground truth labels (Section~\ref{sec:dataandannotation}) at the end of the video. The errors (Section~\ref{sec:metrics}) are then averaged across all points to obtain the final metrics in 2D or 3D. Latency is also calculated alongside the inference for those who participated in the efficiency component of the challenge.}
	\label{fig:challengeexample}
\end{figure*}

In this section, we will first provide a brief clarification of the challenge dataset compared to the original STIR dataset in~\ref{sec:diffs}, followed by a non-exhaustive summary of datasets that we see as useful to tracking in Section~\ref{sec:datasets}.
After which, we will describe the dataset format and annotation protocol for the challenge in Section~\ref{sec:dataandannotation}.
Then, we describe the metrics we calculate as part of the challenge in Section~\ref{sec:metrics}.
We then summarize all submissions received in Section~\ref{sec:submissions}, and their results in Section~\ref{sec:results}.
We provide a discussion of the results and challenge organization in Section~\ref{sec:discussion}, and finally conclude in Section~\ref{sec:conclusion}.
For a high-level overview of the challenge, refer to Fig.~\ref{fig:challengeexample}.

\subsection{STIR Challenge Data (STIRC2024) vs STIR Original (STIROrig)}
\label{sec:diffs}
Here, we will explain the differences between the STIR Challenge 2024 and the STIR dataset.
The original STIR dataset (STIROrig) is a dataset that is publicly released and usable for test, validation, or training (available at: \url{https://ieee-dataport.org/open-access/stir-surgical-tattoos-infrared}).
This dataset is released as a way to validate, test, design, and evaluate algorithms~\cite{schmidtSurgicalTattoosInfrared2024}.
STIROrig remains useful for this exact purpose, in addition to being larger than the challenge dataset.
The STIR Challenge 2024 dataset (STIRC2024) is a similar dataset that was witheld from the initial STIR dataset release in order to enable proper evaluation without the risk of participants fine-tuning or overfitting to already released data.
STIRC2024 has additional filtering and removal of noisy labels, and can be used for fine grained evaluation and testing.

\subsection{Useful Datasets for Tracking in Surgery}
\label{sec:datasets}

At MICCAI 2022, a similar challenge was organized for the same task of tracking tissue~\cite{cartuchoSurgTChallengeBenchmark2024}.
The primary differences are our use of a point-tracking metric~\cite{doerschTAPVidBenchmarkTracking2022}, and increased size and diversity of our data.
In addition, the STIR dataset is not labeled in a temporally dense manner, while the SurgT dataset is labelled per-frame.

There are many other datasets that can benefit tissue tracking.
For a detailed summary of useful datasets in this space, refer to the review~\cite{schmidtTrackingMappingMedical2024a}.
Since that review, additional datasets and meta-datasets have become available.
Here is a brief list of data we recommend looking at.
Meta-MED~\cite{buddTransferringRelativeMonocular2024a} is an assembled meta-dataset.
This dataset is intended to be used for training and evaluating monocular depth models, but would also be useful for self-supervised training of tracker models.
The StereoMIS~\cite{hayozLearningHowRobustly2023} dataset comprises many stereo sequences and could be used for similar purposes.
The SurgVU dataset~\cite{ziaSurgicalVisualUnderstanding2025} also serves as a large (hundreds of hours) repository of single-eye video that could be used for self-supervised training.

\section{Dataset and Annotation}
\label{sec:dataandannotation}

\begin{figure}[t]
\centering
\includegraphics[width=\linewidth]{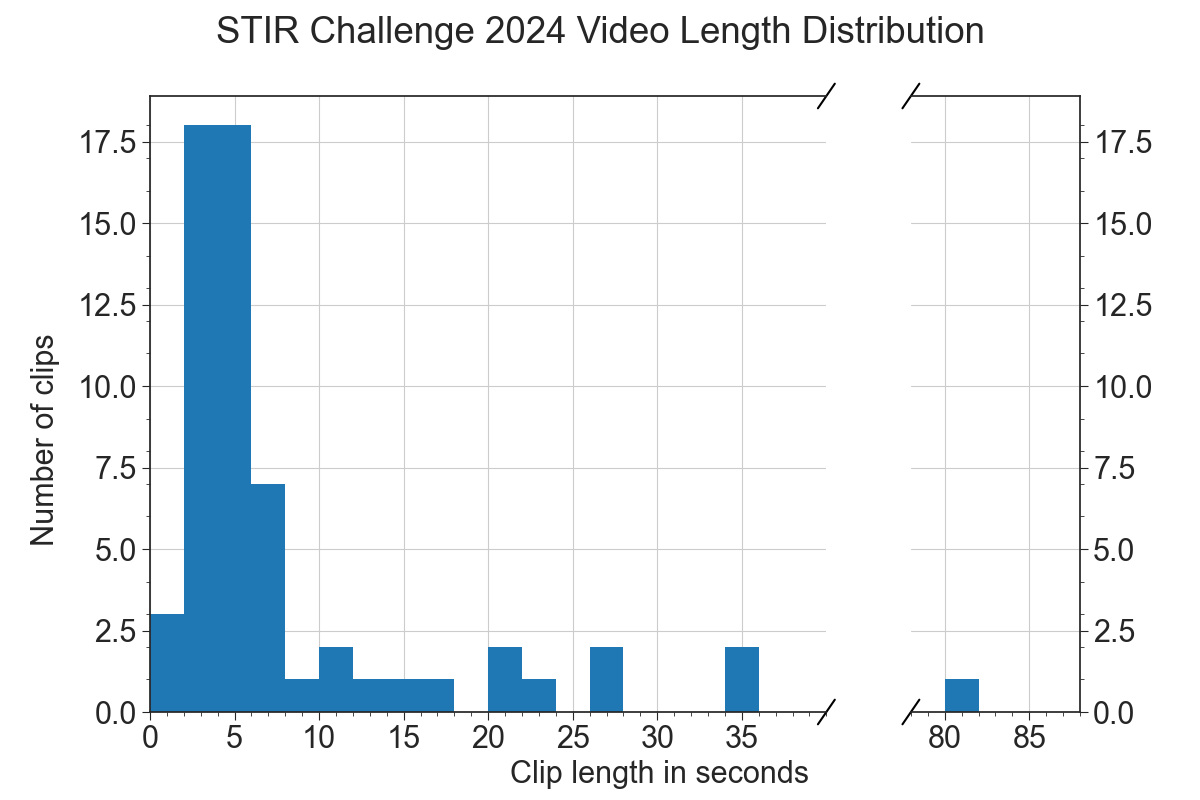}
\caption{Temporal distribution of videos. Most clips lie between 0 and 10 seconds, with a few longer clips \(>20\) seconds. Average clip length is 8.9 seconds.}
\label{fig:cliphistogram}
\end{figure}

\begin{figure}[t]
\centering
\includegraphics[width=\linewidth]{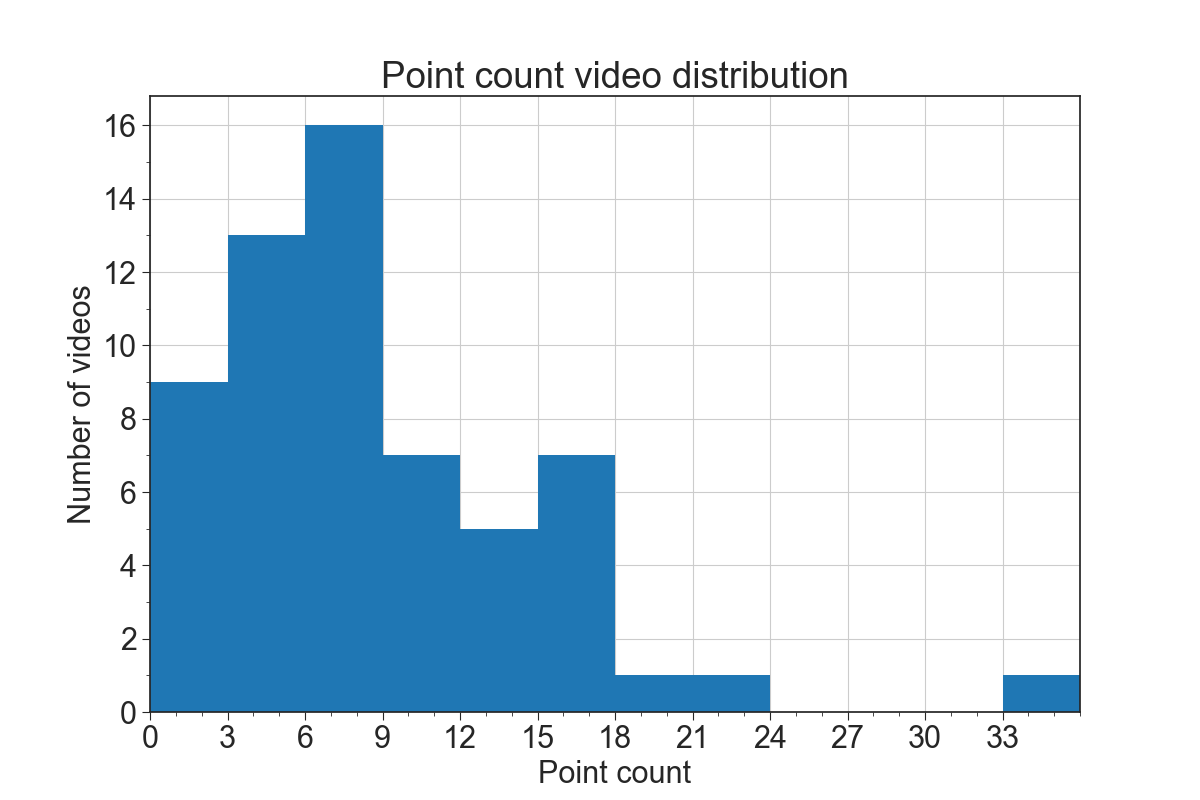}
\caption{Number of labelled points per video. Labels can be seen in Fig.~\ref{fig:startsegs}.}
\label{fig:pointhistogram}
\end{figure}

\begin{figure}[t]
\centering
\includegraphics[width=\linewidth]{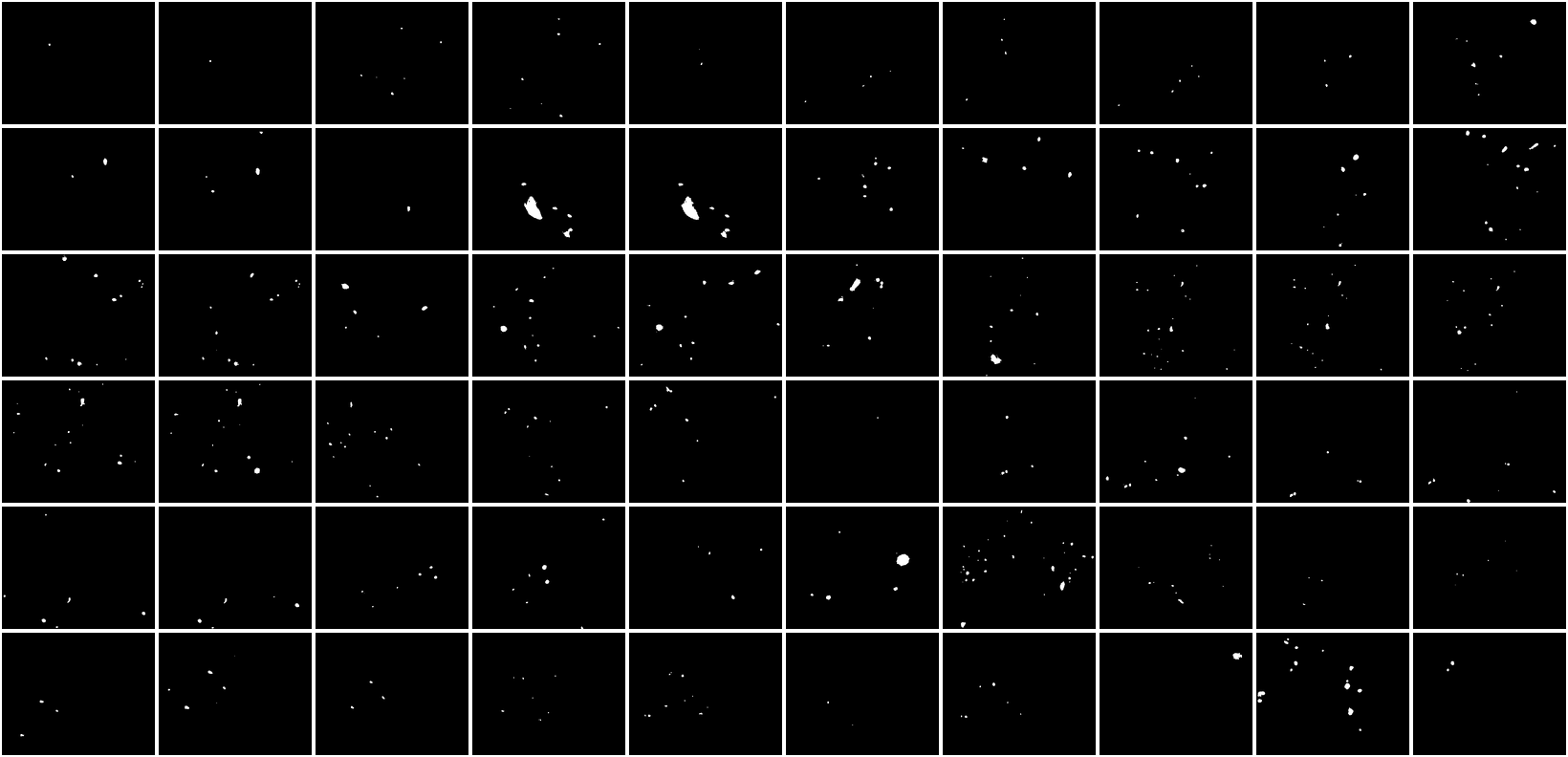}
\caption{Start point labels for all 60 sequences in in the STIR 2024 test dataset. For each sequence, center points are extracted from each segmentation, and passed to each participant's tracker.}
\label{fig:startsegs}
\end{figure}

The STIR Challenge 2024 dataset consist of sets of stereo video clips collected on a da Vinci Xi system.
Each clip consists of a start IR image \(I_s \) and an end IR image \(I_e \), segmentations of the fluorescent ink \( S_s \) and \(S_e \), respectively, and the visible light clip \( V \).
All frames are of size 1280 \(\times\) 1024 pixels.
\(I_s, I_e\) are in Portable Network Graphic (png) format;
\(V\) is the action video in MPEG-4 Part 14 (mp4) format;
\(S_s, S_e\) are binary segmentations of the IR frames (png).
This dataset comprises 60 sequences.
Their average length is 8.9 seconds, with a standard deviation of 12.1 seconds.
The distribution can be seen in Fig.~\ref{fig:cliphistogram}.
For a histogram of points per video refer to Fig.~\ref{fig:pointhistogram}.
No clips longer than 4 minutes are included.
Summary images of the labels can be found in Fig.~\ref{fig:startsegs}.
There are a total of 496 points over the 60 sequences.

\begin{figure*}[t]
	\centering
	\includegraphics[width=\textwidth]{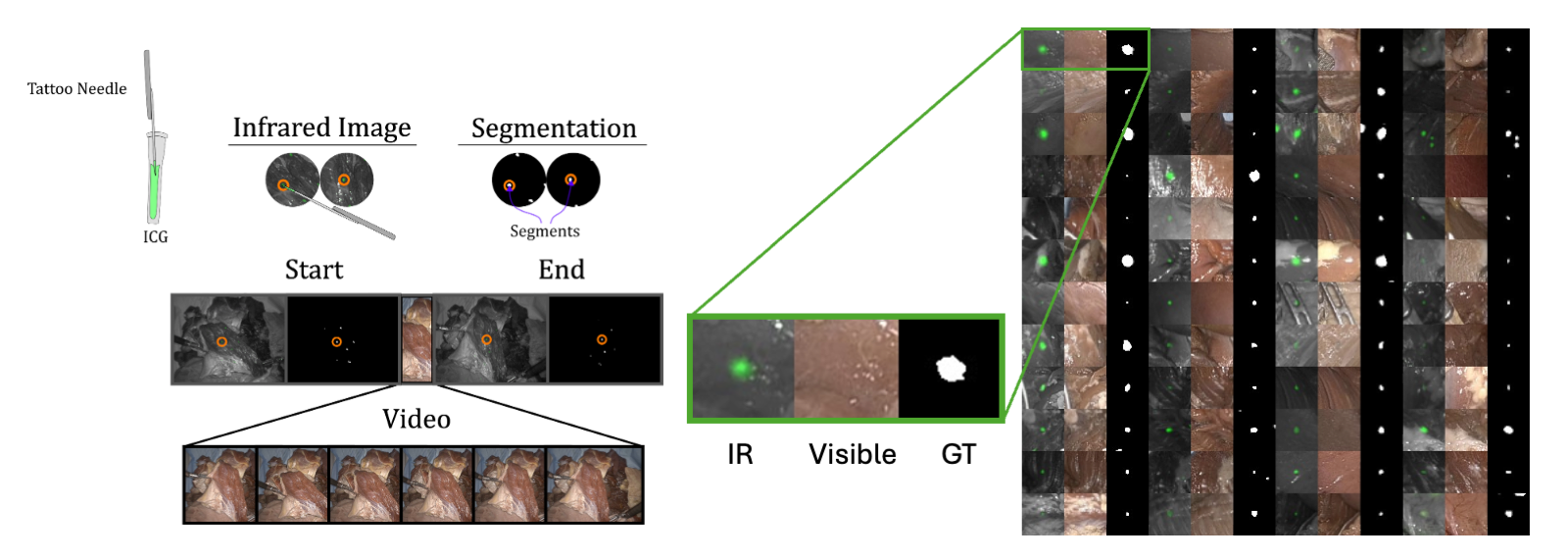}
	\caption{Dataset labels and label creation process. The ground truth is collected by using a tattoo needle to label points at the start and end of video frames. After tattooing is completed, multiple sequences can be collected. For each sequence, the camera captures an image in infrared (ground truth start frame), then switches to white light. Actions are performed under white light, and this video is recorded and saved. Then the camera switches back to IR and captures the end frame which is used as the ground truth for each point's motion. Segments are the binary-thresholded IR images; tattooed regions are shown in white. On the right is a figure showing a set of random triplets with the triplet: (IR image, visible light image, segment/GT image) for each point shown.}
	\label{fig:datalabelling}
\end{figure*}

This dataset was created following the same process as STIROrig~\cite{schmidtSurgicalTattoosInfrared2024}, and more detail can be found there.
The labelling process is visually described in Fig.~\ref{fig:datalabelling}.
At a high level, points are tattooed with indocyanine green (ICG) ink, to create ground truth labels.
The endoscope is switched to fluourescent mode at the start and end of an action to collect the point locations at the start and end of a video.
The video for tracking is recorded in white light, and multiple actions can happen within.
The data comes from porcine subjects for the {\em in vivo} cases, and is a mix of different tissue for the {\em ex vivo} cases.

Segmentations are created by first thresholding the IR-channel of the image.
An opening morphological transformation, which consists of erosion followed by dilation, is applied to reduce noise.
The resulting segments are then verified by ensuring that if a segment appears in the start image that it also appears in the end image.
To annotate, we first evaluate visibility of markers over randomly sampled cases, ensuring the tattoos do not provide features that algorithms could track~\cite{schmidtSurgicalTattoosInfrared2024}.
After this, a user looks through every case and removes noisy segmentation masks that result from specularity.
This filtering helps to reduce label noise.

In order to compute the ground-truth 3D locations, we complete an epipolar search with normalized cross correlation, using the segmented points as candidates.
This enables us to select which segment in the right image corresponds to a given segment in the left image.

The 3D position for a segment is calculated by backprojecting it.
Since the left and right eyes of the endoscope do not have the same principal point, with the left at \(c_x\) and the right at \(c'_x\), we must backproject with this in mind.
We calculate depth using the baseline \(b\), focal length \(f\), and \(c_x, c'_x\) from the calibration along with the point x-location in the left and right image (\(x, x'\)).
The depth, \(z\), is:
\[z = \frac{b*f}{(x - c_x) - (x' - c'_x)}\]

\subsection{Data Format}
\begin{figure*}[t]
	\centering
	\includegraphics[width=\textwidth]{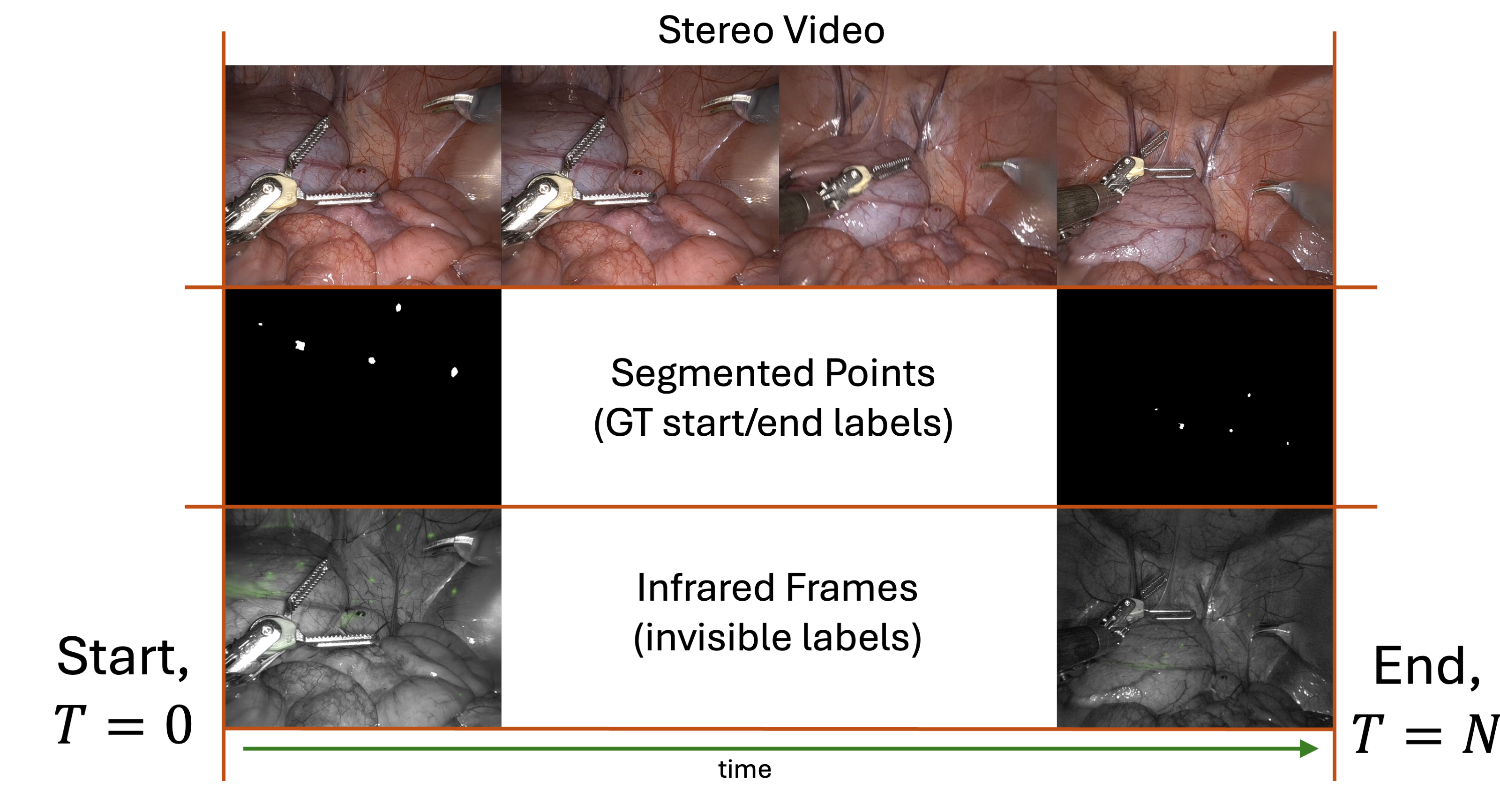}
	\caption{Example {\em in vivo} sequence (session 04, sequence 10) from the STIR Challenge 2024 test set. The start segmentation (middle, left) is converted into a set of points, 5 in this case, that are passed to the algorithm which tracks in the white light stereo video (top). The algorithm results are compared to the ground truth end points (middle, right).}
	\label{fig:dataexample}
\end{figure*}

We summarize the dataset of the STIR Challenge 2024 (STIRC2024) here, noting the format is the same as that for STIROrig~\cite{schmidtSurgicalTattoosInfrared2024}.
STIRC2024 includes a set of 9 collection sessions, named as \textbf{\code{\textless \%02d\,\textgreater}}, \((02, 03, 04, 05, 06, 07, 08, 09, 11)\).
There are 5 {\em in vivo} sessions \((03, 04, 07, 08, 11)\) and 4 {\em ex vivo} sessions \((02, 05, 06, 09)\). Each session includes multiple sequences.
An example sequence for one of the {\em in vivo} cases is shown in Fig.~\ref{fig:dataexample}.

\begin{itemize}
\item[] \code{left}
    \begin{itemize}
        \item[] \code{starticg.png} (Infrared ICG image of start frame)
        \item[] \code{endicg.png} (Infrared ICG image of end frame)
        \item[] \code{segmentation/startim.png}, 
        \item[] \code{segmentation/endim.png} (Filtered and segmented binary versions of ICG start and end image)
        \item[] \code{frames/\textless ms\,\textgreater\_ms.mp4} (video file)
   \end{itemize}
\item[] \code{right}
    \begin{itemize}
        \item[] \code{starticg.png}
        \item[] \code{endim.png}
        \item[] \code{frames/\textless ms\,\textgreater\_ms.mp4} (video file)
    \end{itemize}
\item[] \code{calib.json} Camera calibration parameters (intrinsics, relative stereo pose translation in metres and axis-angle rotation format)
\end{itemize}
The video file names include start and end capture times in milliseconds.


\section{Metrics}
\label{sec:metrics}

In this challenge, we evaluated the submitted algorithms based on two different metrics: accuracy and efficiency.
The accuracy metric is important for clinical verification. The efficiency metric measures the timing latency of the submitted algorithms and evaluates an algorithm's feasibility in running on clinical systems.
For the accuracy metric, we used two categories: 2D trackers, and 3D trackers.

\subsection{Accuracy}

To evaluate accuracy, we use a metric which manages outliers well via calculating accuracy over multiple thresholds.
This metric can be used easily in 2D and 3D.
The metric is \(\delta^{avg}\), introduced in TAP-Vid~\cite{doerschTAPVidBenchmarkTracking2022}, which is a non-medical point tracking challenge.
In TAP-Vid~\cite{doerschTAPVidBenchmarkTracking2022}, the points also have an occlusion score, but here we use data in which the points are unoccluded at the end frame.
In our scenes points can be occluded and reappear during a sequence due to camera movement, instrument-tissue occlusion, or tissue-tissue occlusion.

To calculate the metric, \(\delta^{avg}\), in our case, each algorithm estimates the position for a point (or multiple points) for each frame in a video in a streaming manner.
The final frame location estimate results in a point (or multiple points), \(\hat{p}_{end}\).
The calculated finish points, \(\hat{p}_{end}\), are 2 dimensional for 2D trackers, and 3 dimensional for the 3D trackers.
The accuracy metric is averaged across all points and thresholds, with each point weighted evenly.
To calculate euclidean distance, each point is matched to its nearest point in the end point label.

\begin{align}
\delta^{avg} = \Sigma_{i = i}^{M} \delta^{\mathbf{l}_i}/M\\
\delta^{\mathbf{l}_i} = \Sigma_{\hat{p}_{end}} \mathbbold{1}({d(\hat{p}_{end}, p^{nearest}_{end}) < \mathbf{l}_i}) / N
\end{align}

\(\mathbbold{1}\) is the indicator function, used to count the amount of points under the distance threshold.
\(d()\) is a function to estimate euclidean distance in the dimension of input (2D/3D).
\(N\) is the total number of points across all videos, and \(M\) is the number of thresholds used.
Thus, \(\delta^{\mathbf{l}_i}\) is accuracy at the threshold \(\mathbf{l}_i\).
For 2D, the thresholds are \(\mathbf{l} = [4, 8, 16, 32, 64]\) with units as pixels in the full \(1024 \times 1280\) image.
For 3D, the thresholds are \(\mathbf{l} = [2, 4, 8, 16, 32]\) with units as millimetres.

\subsection{Efficiency}
The efficiency of an algorithm is measured by its computational latency, assessed across all video frames to derive a latency distribution. While the mean latency provides a general efficiency metric, it fails to capture worst-case behavior. In real-world surgical point tracking, predictability depends on worst-case and tail latencies. Therefore, we evaluate efficiency using the 95th and 99th percentile latencies in addition to the mean. The final efficiency score is the average of these three metrics, offering a comprehensive assessment of both real-time and practical performance. A submission is considered for the efficiency category only if the accuracy of the algorithm is above a certain threshold.

\section{Submissions + Baselines}
\label{sec:submissions}

Here, we summarize submissions to the challenge.
The submissions are grouped by those that who were participants for the challenge day (pre-challenge), who were viable for prizes, and those afterwards (post-challenge).
We also provide the results from the baseline methods that we provided on our github page, \url{https://github.com/athaddius/STIRMetrics}.

\subsection{Baselines}
\subsubsection{MFT}
\label{subsubsection-mft}

This is the baseline MFT method~\cite{neoral2024mft}.
This method runs optical flow between a frame and multiple frames at skips into the past.
The algorithm selects its optimal trajectory by selecting the highest certainty unoccluded trajectory.
RAFT is used as the optical flow method in.
To maintain inference efficiency, images are downsampled by a factor of 2 for tracking.
Skip factors are the same as those used in the MFT paper \([-\infty, 1, 2, 4, 8, 16, 32]\)
The occlusion threshold is set as \(0.02\).
After tracking, the locations are scaled up by 2 to get the coordinates in full resolution.

\subsubsection{CSRT}
\label{subsubsection-csrt}
This method uses the Channel and Spatial Reliability tracker (CSRT~\cite{lukezicDiscriminativeCorrelationFilter2017}), initialized with a region of interest around each point in the first frame.
This tracker uses a correlation based adaptive template matching to track a point across a video.
Tracking is performed on half scale images, and upscaled.
The region of interest for each point is a \(29\times 29\) box with its center as the point location.

\subsubsection{RAFT}
\label{subsubsection-raft}
In the RAFT implementation, we use RAFT~\cite{teed2020raft} off-the-shelf to track points from one frame to the next in a streaming manner.
We track on half-scaled images, for efficiency, and multiply the final tracking result by 2 to obtain full-resolution results.
RAFT internally iterates multiple times, refining estimates with iteration.
We use 12 iterations in inference.

\subsubsection{RAFT + RAFT Stereo (3D)}
\label{subsubsection-raft3D}
This is the 3D baseline, which uses RAFT to estimate flow from one frame to the next in the left eye, and finds the 3D position by backprojecting the points using the disparity estimate alongside the camera calibration.
The disparity estimate is calculated using RAFT-Stereo~\cite{lipson2021raft}.

\subsubsection{Control}
\label{subsubsection-control}

The control method estimates 0 motion for every point.
This provides a minimum bound of accuracy which is useful for debugging and static scenes.
This also allows the organisers and participants to ensure the challenge methods use the correct data.
The control method runs alongside submissions.
During the challenge, this served as a useful sanity check.

\subsection{Challenge Day}

Four teams were able to submit in time for challenge day.

\subsubsection{Team ICVS\_2AI}
\begin{figure}[t]
\centering
\includegraphics[width=\linewidth]{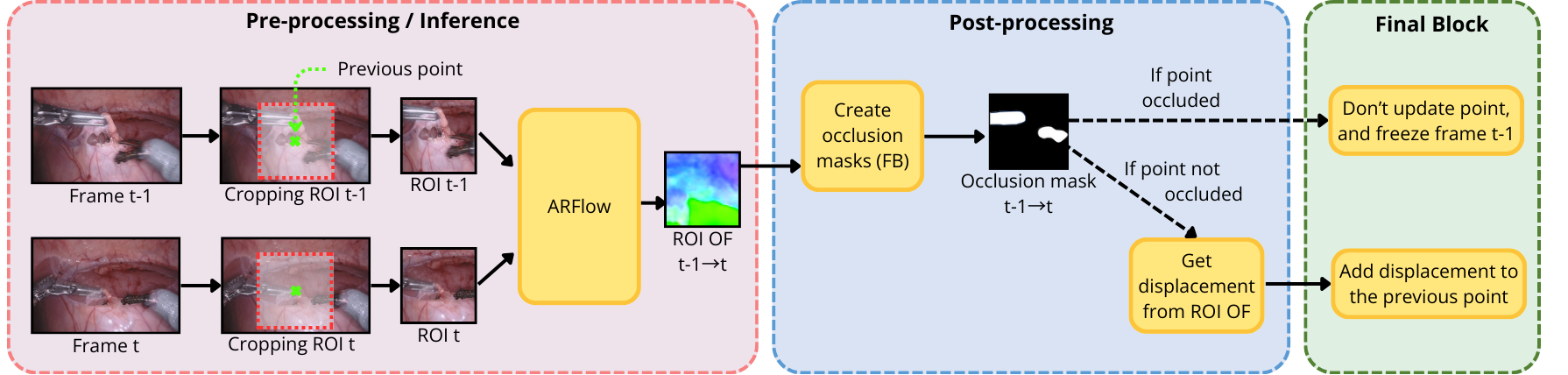}
\caption{Overview of the solution provided by team ICVS\_2AI. The post-processing block in the baseline method utilized forward-backward occlusion masks.}
\label{icvs-2ai-main-fig}
\end{figure}

Team ICVS\_2AI proposes an occlusion-aware optical flow-based solution.
Targeting to tackle the problem of labeled-data-scarcity in surgical domain, the method employs ARFlow as the optical flow model trained on the SurgT dataset in a self-supervised fashion following previous work~\cite{silva2024evaluating}. The proposed architecture, which is depicted in Fig.~\ref{icvs-2ai-main-fig}, first crops a $512\times512$ region around the location of the last track location, then estimates the optical flow between the source target frames. To prioritize runtime performance, for the efficiency component, the point tracks are only computed between previous frame $t-1$ and current frame $t$; however, in the 2D accuracy challenge, frame $t-2$ is also used as a secondary source frame and the final point track is estimated as the weighted average of the two computations. For the 3D challenge, the estimated 2D point track is lifted to the 3D space by stereo-depth computation applying the flow-model between left-right image pairs.

\subsubsection{Team MedTrack}
\begin{figure}[t]
\centering
\includegraphics[width=\linewidth]{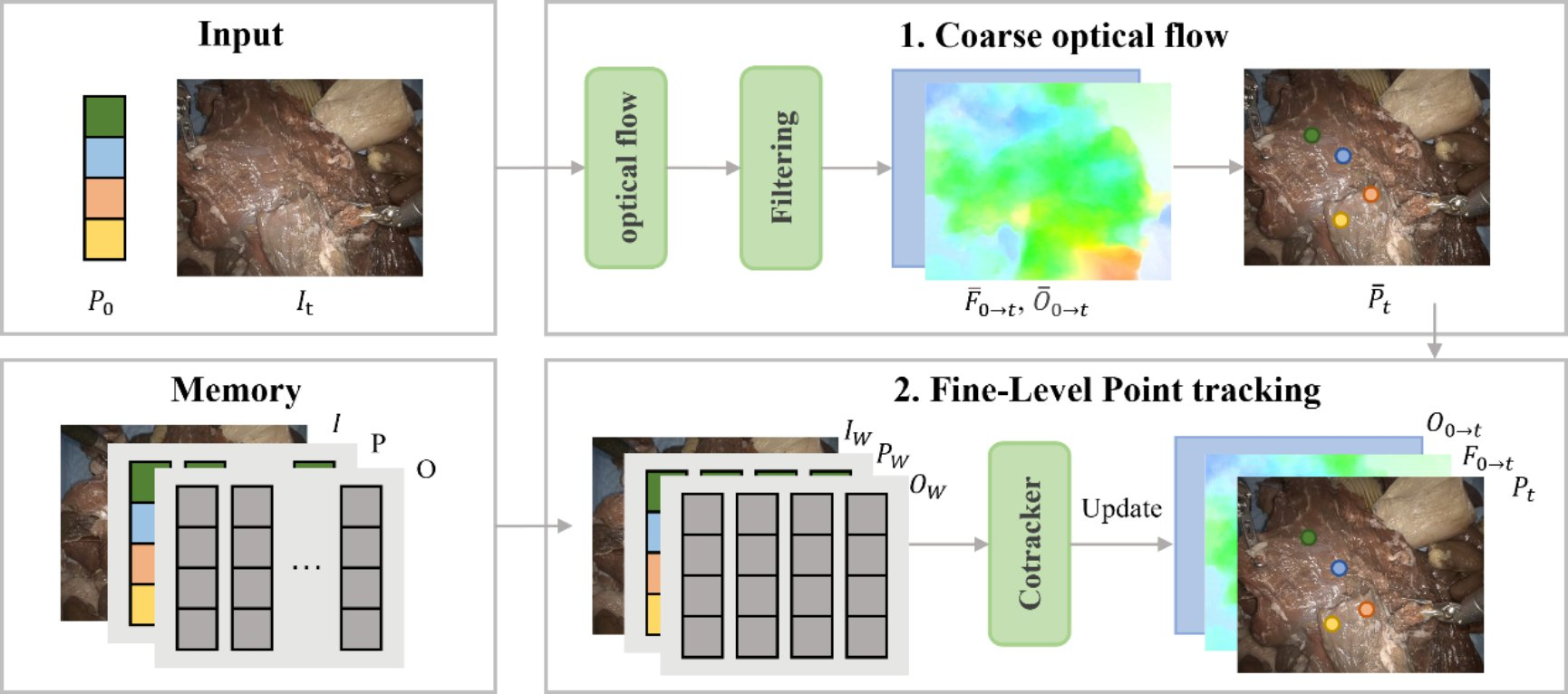}
\caption{Overview of the solution provided by team MedTrack: DMPTracking. First, it uses coarse optical flow to provide an initial tracking position Pt for the query points P0, and then refines the tracking results using multi-frame information.}
\label{medtrack-main-fig}
\end{figure}

Team MedTrack proposes a two-step, hierarchical, long-term tracking method called Dynamic Multi-Frame Point Tracking (DMPTracking) depicted in Fig.~\ref{medtrack-main-fig}.
In the first step, DMPTracking employs an MFT-based~\cite{neoral2024mft} approach to estimate point tracks and their visibilities at a coarse level.
Following MFT's~\cite{neoral2024mft} original structure, as explained in further detail in Section ~\ref{subsubsection-mft}, it computes the optical flow, uncertainty, and visibility scores between the current frame and a set of geometrically distanced past, template, frames and estimates the track of a point through a temporal chaining mechanism.
However, this template-based optical flow mechanism can provide erroneous solutions if there are large brightness changes between the templates and the current frame.
DMPTracking utilizes a filtering mechanism that checks the magnitude of the spatial gradient of the optical flow map and replaces the values that are above a threshold with interpolated values from the local neighborhood.

In the second step, DMPTracking utilizes a CoTracker-based~\cite{karaev2025cotracker} structure to refine the coarse-solution.
As an end-to-end learning based approach, CoTracker~\cite{karaev2025cotracker} processes a video sequence in chunks with a  window size of 8 frames and in each step slides the window by a stride of 4 frames.
In certain cases, such as occlusions spanning multiple windows or long sequences, this window-based processing can cause failures in tracking.
To prevent this, the proposed architecture alters the structure of the sliding window, and instead fixes the first and last frames of it to be, respectively, the initial frame of the entire sequence and the current frame.
The remaining 6 frames are chosen from previously observed frames.

\subsubsection{Team Jmees}
Team Jmees employs MFT~\cite{neoral2024mft} for both 2D and 3D tracking tasks.
For 2D tracking, the model processes input frames at \( 1/4 \) of their original size to enhance computational efficiency.
In the 3D tracking task, points are tracked independently in the left and right frames and subsequently lifted into 3D space using the disparity computed between them.

\subsubsection{Team CCG\_DGIST}

\begin{figure}[t]
\centering
\includegraphics[width=\linewidth]{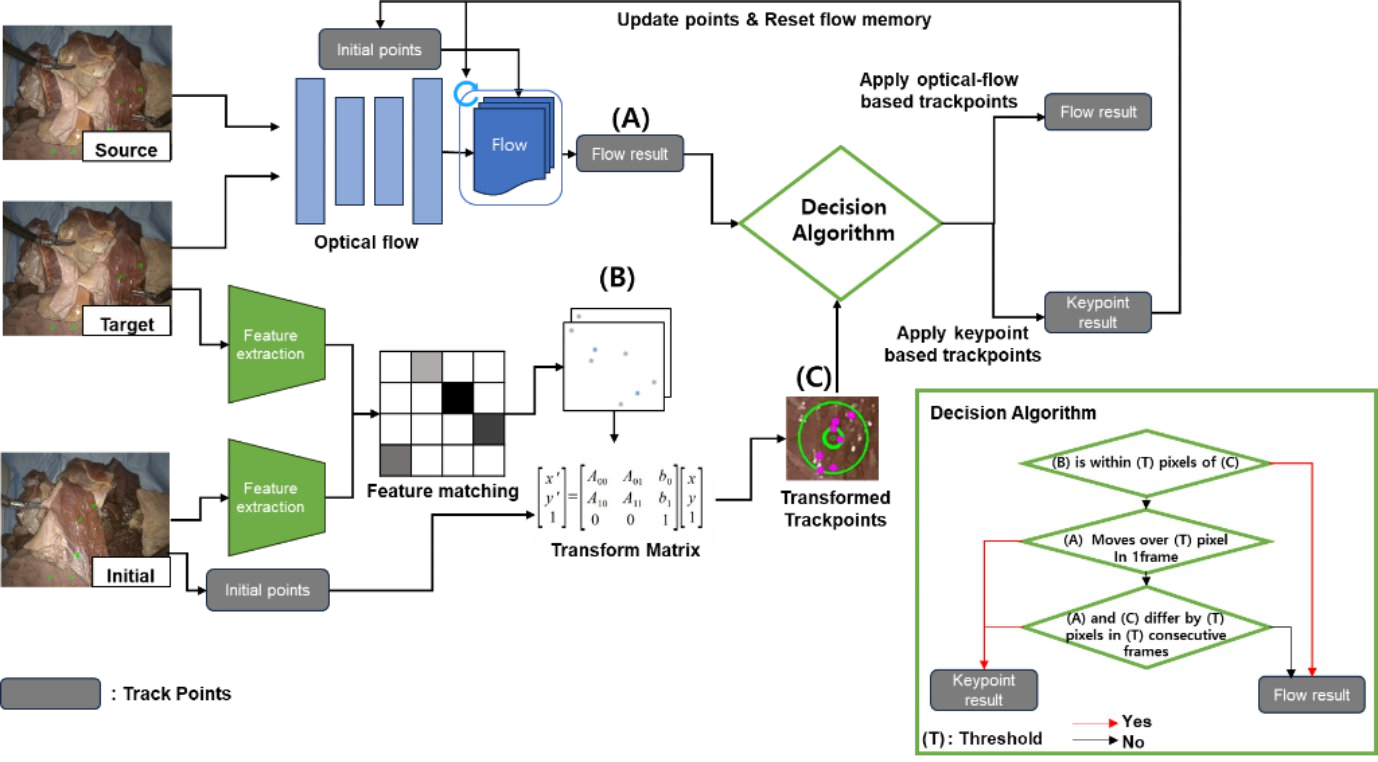}
\caption{Overview of the solution provided by team CCG\_DGIST that combines sparse feature matching and dense optical flow estimation.}
\label{ccg-dgist-main-fig}
\end{figure}

Team CCG\_DGIST proposes a joint sparse keypoint matching and dense optical approach, depicted in Fig.~\ref{ccg-dgist-main-fig}.
The optical flow model consecutively processes frames to estimate and store the point tracks.
Concurrently, the sparse keypoint matching method is utilized to estimate homographic transformations of tracked points between the initial and target frames.
The final track update is selected based on a decision algorithm.
In more detail, the optical flow based decision is selected if a matched keypoint exists within a certain threshold of the optical flow estimation.
In the case of detection of a drift of the optical flow estimation with respect to the keypoint-based estimation, the latter's estimation is selected as a correction.
Finally, if the displacement between the keypoint and the optical flow based method is above a certain threshold the keypoint-based estimation is selected.

\subsection{Post-Challenge Day}
\subsubsection{Team CTUPrague}

Team CTUPrague participates with their MFT~\cite{neoral2024mft}-extending method, MFTIQ~\cite{serych2024mftiq}.
Similar to MFT, MFTIQ uses the optical-flow chaining structure and replaces the implicit occlusion and uncertainty estimation of the optical flow model with an independent network that aggregates the warped feature maps with a feature similarity cost map to compute the quality and occlusion scores.
This renders further adaptability of the architecture with various flow estimation backends.
In this challenge, their submissions utilize two separate flow models: SEA-RAFT~\cite{wang2024sea} and ROMA~\cite{edstedt2024roma}.

\subsubsection{Team UBC\_RCL}

Team UBC\_RCL participates with A-MFST~\cite{chen2024mfst}.
Extending the flow-chaining architecture of MFT~\cite{neoral2024mft}, A-MFST replaces the RAFT~\cite{teed2020raft}-based optical flow backend with SENDD~\cite{schmidt2023sendd} and the certainty and occlusion estimation with backward-forward flow consistency.
This consistency score is also used for frame selection to prioritize more reliable template curation.
Instead of storing input images as templates, as in MFT~\cite{neoral2024mft}, A-MFST caches previously computed features to reduce redundant computations in the flow-estimation. 

\subsubsection{Team SRV}

Team SRV participates with a frame-to-frame tracking method, SEA-RAFT~\cite{wang2024sea}.
Focusing on efficiency improvements, SEA-RAFT extends over the prior work RAFT~\cite{teed2020raft} and introduces a more efficient architecture.
Coupled with a more robust training strategy, it additionally shows accuracy improvements.

\subsubsection{Team CUHK}
\begin{figure}[t]
\centering
\includegraphics[width=\linewidth]{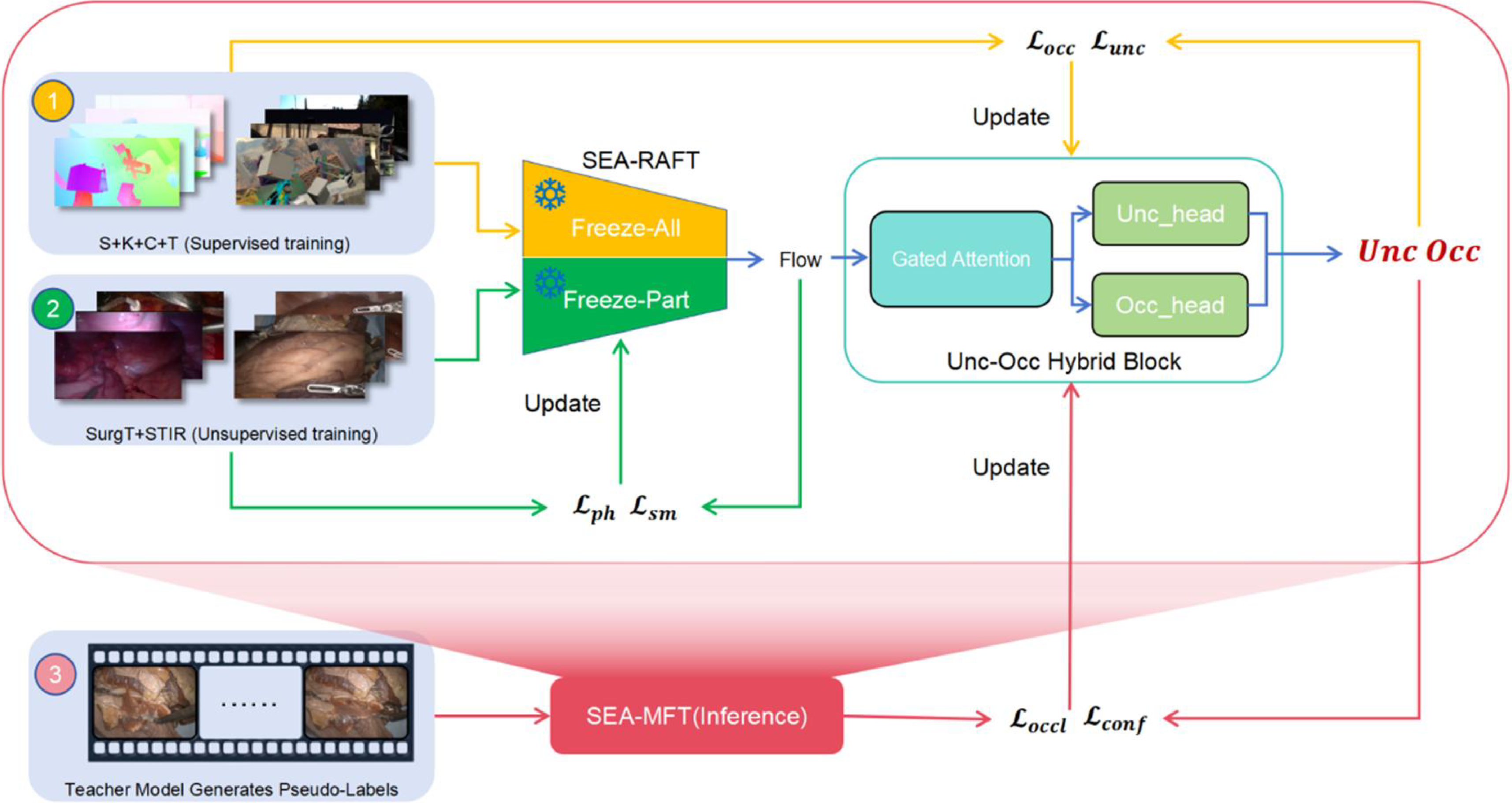}
\caption{Overview of the solution provided by team CUHK: TAP-Endo. It consists of a three step training structure: (1) supervised training of the uncertainty and occlusion block; (2) fine-tuning of the optical flow model using endoscopic data in a self-supervised fashion; (3) fine-tuning of the uncertainty and occlusion block leveraging pseudo-labels acquired using a set of state-of-the-art point  tracking methods.}
\label{cuhk-main-fig}
\end{figure}
Team CUHK participates with TAP-Endo, a semi-supervised method extending MFT~\cite{neoral2024mft}.
The proposed architecture utilizes SEA-RAFT~\cite{wang2024sea} to replace MFT's original optical flow component.
The architecture uses a three step training strategy to improve generaliziblity to endoscopic scenes.
Their method is shown in Fig.~\ref{cuhk-main-fig}.
In the first step, a pretrained SEA-RAFT with frozen weights are appended with a gated attention module and trained in supervised fashion on synthetic datasets to predict the uncertainty and occlusion.
In the second step, a part of the SEA-RAFT backbone is unfrozen and refined using endoscopic datasets~\cite{cartuchoSurgTChallengeBenchmark2024, schmidtSurgicalTattoosInfrared2024} in a self-supervised fashion~\cite{liu2020learning}.
In the final step, the occlusion and uncertainty heads are fine-tuned using pseudo-labels generated by a set of state-of-the-art point tracking methods~\cite{karaev2024cotracker3, neoral2024mft, li2024taptr, doersch2023tapir}.

\section{Results}
\label{sec:results}

This section will detail and analyze the results for each challenge participant.

\subsection{Accuracy}

\begin{table}[t]
\caption{2D tracking accuracy comparison. $\delta^{\mathbf{l}_i}$ indicates the tracking accuracy where $\mathbf{l}_i$ is the threshold defined in pixels.}
\label{tab:accuracy-2d}
\begin{tabularx}{\columnwidth}{X c c c c c c c}
    \toprule
    \multirow{2}{*}{\textbf{Method}} & \multicolumn{6}{c}{$\delta^{\mathbf{l}_i}\uparrow$}\\
    \cmidrule{2-7}
    & $\mathbf{l}_i=4$ & $\mathbf{l}_i=8$ & $\mathbf{l}_i=16$ & $\mathbf{l}_i=32$ & $\mathbf{l}_i=64$ & Avg. \\
    \midrule
    \multicolumn{7}{X}{\textbf{Baselines}} \\
    \midrule
    RAFT & 07.26 & 19.56 & 39.92 & 64.92 & 80.44 & 42.42 \\
    CSRT &  22.78 & 47.38 & 67.14 & 74.80 &  81.05 & 58.63 \\
    MFT & 42.54 & 69.36 & 86.49 & 93.35 & 96.37 & 77.62 \\
    \midrule
    \multicolumn{7}{X}{\textbf{MICCAI\newline Submissions}} \\
    \midrule
    ICVS\_2AI & 25.40 & 51.01 & 74.19 & 88.71 & 92.54 & 66.37 \\
    JMEES & 26.00 & 54.44 & 77.42 & 91.73 & 95.36 & 68.99  \\
    CCG\_DGIST & 36.09 & 63.51 & 83.87 & 90.93 & 94.36 & 73.75 \\
    MedTrack & 38.91 & 67.54 & 86.69 & 93.15 & 95.97 & 76.45 \\
    \midrule
    \multicolumn{7}{X}{\textbf{Post\newline MICCAI\newline Submissions}} \\
    \midrule
    SRV & 07.86 & 18.15 & 31.65 & 47.78 & 66.73 & 34.44 \\
    UBC\_RCL & 20.57 & 42.34 & 66.73 & 77.42 & 85.89 & 58.59 \\
    MFTIQ-SEARAFT & 42.34 & 68.95 & 85.89 & 92.14 & 94.76 & 76.82 \\
    MFTIQ-ROMA & 44.36 & 69.56 & 85.89 & 91.13 & 95.16 & 77.22 \\
    CUHK & 40.93 & 68.95 & 87.50 & 93.15 & 96.37 & 77.38 \\
    \bottomrule
\end{tabularx}
\end{table}

\begin{table}[t!]
\caption{3D tracking accuracy comparison. $\delta^{\mathbf{l}_i}$ indicates the tracking accuracy where $\mathbf{l}_i$ is the threshold defined in pixels.}
\label{tab:accuracy-3d}
\begin{tabularx}{\columnwidth}{X c c c c c c c}
    \toprule
    \multirow{2}{*}{\textbf{Method}} & \multicolumn{5}{c}{$\delta^{\mathbf{l}_i}\uparrow$}\\
    \cmidrule{2-7}
    & $\mathbf{l}_i=2$ & $\mathbf{l}_i=4$ & $\mathbf{l}_i=8$ & $\mathbf{l}_i=16$ & $\mathbf{l}_i=32$ & Avg. \\
    \midrule
    \multicolumn{7}{X}{\textbf{Baselines}} \\
    \midrule
    RAFT-Stereo & 13.94 & 36.16 & 60.40 & 79.80 & 91.51 & 56.36\\
    \midrule
    \multicolumn{7}{X}{\textbf{MICCAI Submissions}} \\
    \midrule
    JMEES & 25.70 & 45.40 & 65.31 & 81.16 & 91.01 & 61.71 \\
    ICVS\_2AI & 27.88 & 55.15 & 75.96 & 91.11 & 97.58 & 69.54 \\
    \bottomrule
\end{tabularx}
\end{table}

For the 2D methods, Table~\ref{tab:accuracy-2d} provides the overarching summary.
Of the ranked challenge submissions (non-post-challenge), Team Medtrack came in first, with a \(\delta^{avg} = 76.45\).
Team CCG\_DGIST was second with a \(\delta^{avg} = 73.75\), and Team Jmees came in third with a \(\delta^{avg} = 68.99\).
The overall best performing method was MFT~\cite{neoral2024mft}.

For the 3D methods, refer to the summary Table~\ref{tab:accuracy-3d}.
Of the ranked challenge submissions, Team ICVS\_2AI came in first with a \(\delta^{avg} = 69.54\), and Team Jmees came in second with a \(\delta^{avg} = 61.71\).

\subsection{Efficiency}
\begin{figure}[t]
\centering
\includegraphics[width=0.9\linewidth]{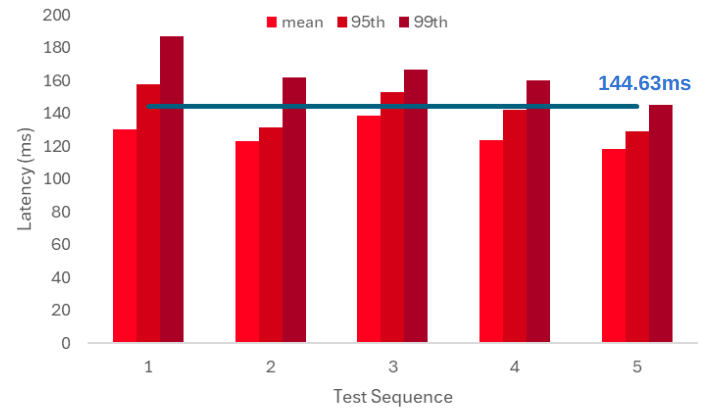}
\caption{Efficiency Result of ICVS\_2AI's method.}
\label{fig:icvs-2ai-efficiency}
\end{figure}

In this section we summarize the efficiency results.
Only the ICVS\_2AI team participated in the efficiency component. To assess their algorithm’s efficiency, we evaluated selected test cases from our dataset, measuring the mean, 95th percentile, and 99th percentile latencies on an NVIDIA A100 GPU. The final latency score averaged across our test cases was 144.63 ms, corresponding to 7 FPS, which is within the acceptable range for many surgical point-tracking applications. Fig.~\ref{fig:icvs-2ai-efficiency} shows the latencies of the ICVS\_2AI submission for 5 test sequences.
The maximum latency remains below 200 ms, a critical threshold in this inaugural efficiency evaluation in this year's challenge. Broader participation in the efficiency component would provide deeper insights into the deployability of competing algorithms.

\section{Discussion}
\label{sec:discussion}
In this section, we will summarize the numerical results from the challenge in Section~\ref{sec:accandeff}, by takeaways from organizing the challenge in Section~\ref{sec:challengelearning}, and finish with suggestions for future algorithms and areas we believe could be useful in Section~\ref{sec:algodirections}.

\subsection{Accuracy and Efficiency}
\label{sec:accandeff}

Here we discuss the 2D results, followed by the 3D results and efficiency.

The most accurate method in the 2D component of the challenge was the baseline MFT~\cite{neoral2024mft} method.
In terms of best fine-grained performance (at \(\delta = 4\)), the baseline MFT also achieved the highest accuracy.
This could be due to the intrinsic memory of MFT which allows it to search back to the first frame.
Methods that encountered difficulty over sequences in this challenge were those which do not support longer window recovery (RAFT (baseline), SEA-RAFT (Team SRV)).
Frame-wise methods (or methods with short windows) still attained reasonable performance when integrating occlusion masks (Team ICVS\_2AI).

Regarding long-term methods, CCG\_DGIST addressed long term tracking by having two branches to decide between an optical flow method and a homography transformation to re-detect points after they may be occluded.
To deal with longer occlusion, Team Medtrack uses MFT along with CoTracker.
They alter the sliding window of CoTracker to include the initial frame to help performance under long time spans.
In Medtrack's submission, MFT is used to calculate initial positions. The optical flow for MFT is also filtered to remove outliers in non-smooth regions using the spatial gradients of the flow map.
The candidates from MFT are then used as initial points to seed CoTracker.
Team Medtrack attained the second highest score among the methods submitted for 2D accuracy tracking.
The TAP-Endo method submitted by Team CUHK extends SEA-RAFT~\cite{wang2024sea} with an occlusion and uncertainty module and fine-tunes it using a semi-supervised strategy.
They integrate their method into the MFT architecture for robust point tracking with a longer temporal context allowing them to achieve the highest accuracy of all the submissions for the task of 2D tracking.

Interestingly, Team MFTIQ~\cite{serych2024mftiq} (ROMA/SEA-RAFT) and TAP-Endo both achieved competitive performance but did not surpass the baseline MFT.
TAP-Endo (Team CUHK) employed a domain-adaptation strategy, yet still fell short of outperforming the baseline.
This outcome may indicate that existing backbone optical flow architectures and refinement strategies, generalize suboptimally or inconsistently to the unique challenges posed by the endoscopic domain.
We expect further exploration and optimization in this area could significantly enhance performance and robustness in future.

Notably, most of these methods do not train a long-term tracking or occlusion management method.
ICVS trains an optical flow on SurgT, and UBC\_RCL trains an optical flow method on image pairs.
Although Medtrack uses CoTracker with the start frame as the first frame in each sliding window, the CoTracker model is not trained under these surgical scenarios.
It could be expected that more focus on the occlusion management and re-detection could improve performance in future submissions.

In terms of 3D methods, Team ICVS\_2AI had the most accurate submission.
Team ICVS\_2AI tracks in 2D and uses this 2D track along with estimated depth to obtain 3D locations.
Team Jmees tracks a point in each eye, and uses the disparity between these points to backproject.
In terms of future work, a method could be envisioned which either: tracks directly in 3D, or communicates between both the left and right tracks via a transformer-like model or a simpler classical consensus filtering.
In the future, methods that filter or use both frames should be able produce better 3D results in addition to improving tracking in the 2D frame space.
We believe this since more information is available in the stereo data.
As a simple example, a point may be occluded in one eye but not in the other.

Finally, for the efficiency component, although there was only one submission, the emphasis on efficiency in the submissions helped to ensure that the methods are clinically feasible.
This component alongside the constraint that methods must run in a streaming manner, in which they are unable to see future frames, helps to focus the challenge on algorithms that could be usable in a surgical system.

\subsection{Challenge Takeaways}
\label{sec:challengelearning}

We have a few takeaways from the challenge organization as a whole, and we will summarize them here.
For future iterations, we will make the docker submission process more clear in order to better enable quick evaluation without having to debug submissions with challenge teams.
Some teams ran into memory issues with validation on the STIROrig dataset since the dataloader we provide loads videos fully into memory.
Providing a simpler and more lightweight evaluation code structure should help to fix this.

In terms of 3D participation and efficiency, we saw lower participation, and believe this is due to the ease of use for implementing methods within these frameworks.
In the future, we will look to provide more detailed documentation for every component of the challenge.

\subsection{Algorithmic Directions}
\label{sec:algodirections}

For future methods, here is a brief list of ideas that can be focused on:
\begin{itemize}
    \item Using pseudolabelling, synthetic data augmentation, student teacher models applied to surgical scenarios (ie. CoTracker3~\cite{karaev2024cotracker3}, Bootstap~\cite{doersch2024bootstap})
    \item Pretraining models to use surgical features (Masked-autoencoders, self-supervision, etc.)
    \item Using stereo data to improve tracking, rather than tracking in a single eye (left/right).
    \item Training efficient models for long-term tracking, and relocalization.
\end{itemize}

\section{Conclusion}
\label{sec:conclusion}

In this paper, we summarized and discussed the design, data, participation, and results from the 2024 STIR Challenge, which was organized as a part of EndoVis at MICCAI 2024.
We expect this challenge, and the publicly released test dataset will serve as a high-quality resource for methods to test, compare, and iterate on algorithms for tissue tracking and other applications.
The field of image guidance in surgery, and many other applications depend on accurate methods, and see this work as a key step in continuing to enable surgical applications.

\bibliographystyle{splncs04}
\bibliography{reflist}

\end{document}